\documentclass{article}

\usepackage{microtype}
\usepackage{graphicx}
\usepackage{subfigure}
\usepackage{booktabs} 
\usepackage{hyperref}
\usepackage{tikz} 
\usepackage{tikzducks} 
\usepackage{multirow}
\usepackage{tabularx}
\usepackage{bbding}

\usepackage[accepted]{icml2024}

\usepackage{amsmath}
\usepackage{amssymb}
\usepackage{mathtools}
\usepackage{amsthm}
\usepackage{listings}

\usepackage[capitalize,noabbrev]{cleveref}

\theoremstyle{plain}

\theoremstyle{definition}

\theoremstyle{remark}

\newcommand{\eg}{\emph{e.g.}}
\newcommand{\ie}{\emph{i.e.}}

\usepackage[textsize=tiny]{todonotes}

\icmltitlerunning{GeoReasoner: Geo-localization with Reasoning in Street Views using a Large Vision-Language Model}

\begin{document}

\twocolumn[
\icmltitle{GeoReasoner: Geo-localization with Reasoning in Street Views \\using a Large Vision-Language Model}




\begin{icmlauthorlist}
\icmlauthor{Ling Li}{sch1}
\icmlauthor{Yu Ye}{sch2}
\icmlauthor{Yao Zhou}{sch3}
\icmlauthor{Bingchuan Jiang}{sch4}
\icmlauthor{Wei Zeng}{sch1,sch5}
\end{icmlauthorlist}

\icmlaffiliation{sch1}{The Hong Kong University of Science and Technology (Guangzhou)}
\icmlaffiliation{sch2}{Tongji University}
\icmlaffiliation{sch3}{Independent Researcher}
\icmlaffiliation{sch4}{Information Engineering University}
\icmlaffiliation{sch5}{The Hong Kong University of Science and Technology}

\icmlcorrespondingauthor{Bingchuan Jiang}{jbc021@163.com}

\icmlkeywords{Machine Learning, ICML}

\vskip 0.3in
]

\printAffiliationsAndNotice{}  

\begin{abstract}
This work tackles the problem of geo-localization with a new paradigm using a large vision-language model (LVLM) augmented with human inference knowledge.
A primary challenge here is the scarcity of data for training the LVLM - existing street-view datasets often contain numerous low-quality images lacking visual clues, and lack any reasoning inference.
To address the data-quality issue, we devise a CLIP-based network to quantify the degree of street-view images being locatable, leading to the creation of a new dataset comprising highly locatable street views.
To enhance reasoning inference, we integrate external knowledge obtained from real geo-localization games, tapping into valuable human inference capabilities.
The data are utilized to train \emph{GeoReasoner}, which undergoes fine-tuning through dedicated reasoning and location-tuning stages.
Qualitative and quantitative evaluations illustrate that \emph{GeoReasoner} outperforms counterpart LVLMs by more than 25\% at country-level and 38\% at city-level geo-localization tasks, and surpasses StreetCLIP performance while requiring fewer training resources.
The data and code are available at \href{https://github.com/lingli1996/GeoReasoner}{https://github.com/lingli1996/GeoReasoner}.

\end{abstract}
\section{Introduction}
\begin{figure}[t]
  \begin{center}
  \centerline{\includegraphics[width=0.99\columnwidth]{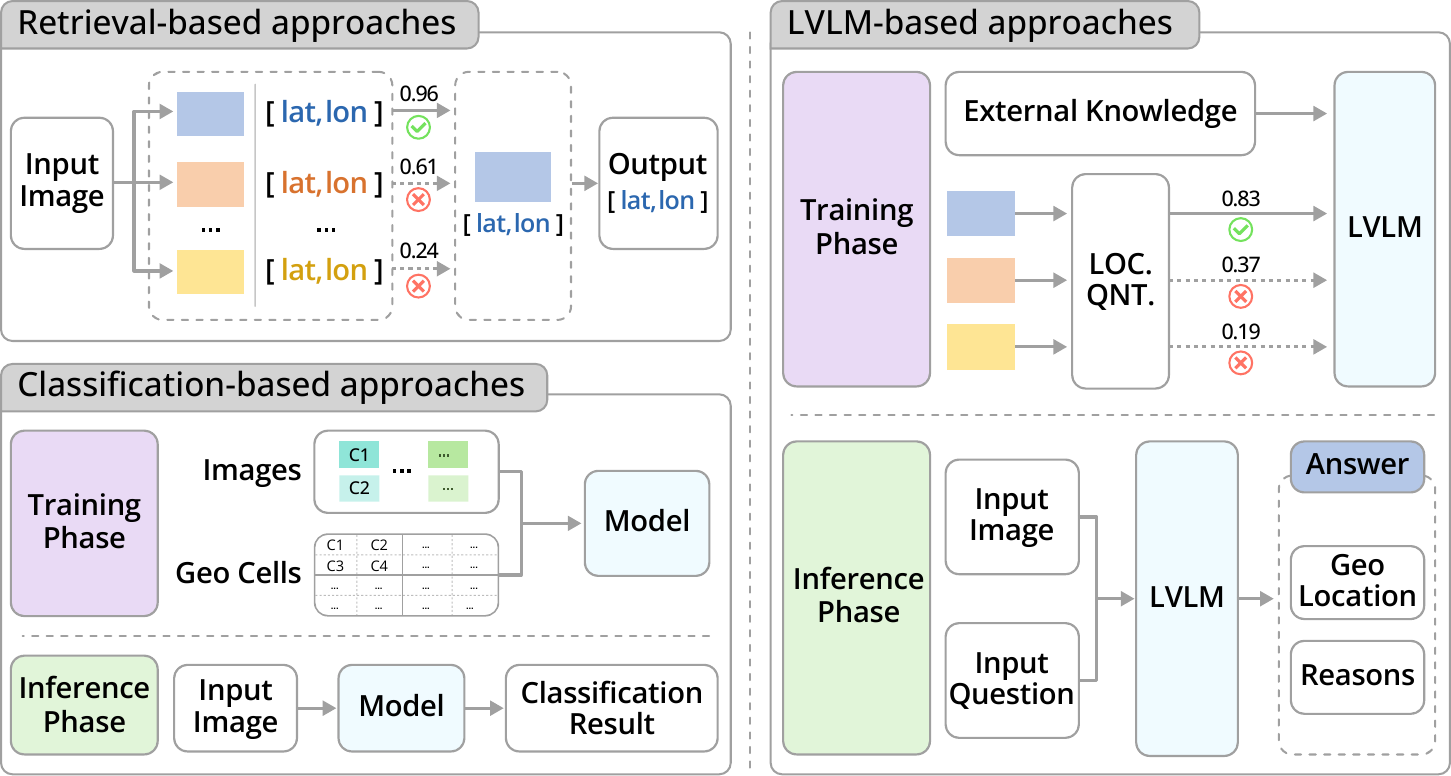}}
  \vspace{-3mm}
  \caption{Different paradigms in existing and the proposed geo-localization approaches: retrieval-based (left-top), classification-based (left-bottom), and our LVLM-based (right).}
  \vspace{-6mm}
  \label{fig:intro}
  \end{center}
\end{figure}

Street-view geo-localization seeks to predict geographical locations for the given street-view images.
The significance of street-view geo-localization is evident in a variety of applications, spanning social studies~\cite{ye_2019_visual}, urban planning~\cite{shen_2017_streetvizor}, and navigation~\cite{chalvatzaras2022}. 
As shown in Figure~\ref{fig:intro} (left), existing frameworks for street-view geo-localization can be mainly divided into two categories:
\emph{retrieval-based} and \emph{classification-based}.
\emph{Retrieval-based} approaches entail identifying the most similar image within a geo-tagged image gallery and returning the corresponding geographical location~\cite{zhu2022, lin2022, zhang2023}.
However, the methods rely on the diversity and comprehensiveness of the geo-tagged image gallery, which can be challenging to curate. 
Alternatively, \emph{classification-based} approaches partition the Earth's surface into distinct regions and assign the input image to a specific region~\cite{clark2023, pramanick2022, muller2018, seo2018, weyand2016}.
While these methods leverage shared visual features within a single region, they may neglect valuable semantic information (\eg, signboard texts) crucial for geo-localization.
More importantly, these classification methods often operate as black-box models, lacking reasoning capabilities for users to interpret.

Achieving street view-based geo-localization with reasoning capability poses a considerable challenge.
This study introduces a new paradigm that facilitates geo-localization with reasoning capability for street-view images, as depicted in Figure~\ref{fig:intro}(right).
The paradigm leverages an LVLM for its excellent capability in handling multi-modal visual and textual inputs and incorporates external knowledge learned from various online games for the reasoning procedure.
Specifically, we introduce the concept of \emph{locatability} as a metric to quantify the degree of locatability in street-view images.
On this basis, we devise a CLIP-based visual-text pairing network to match large-scale Google Street View (GSV) images with 3K finely reasoned text-image pairs from online games, to tackle the challenge of the absence of a high-quality street-view dataset.
The process filters through over 70K GSV images with geo-tags, all of which exhibit a high degree of locatability.

Next, we construct an LVLM model, named \emph{GeoReasoner}, to overcome the difficulty of integrating reasoning capability in geo-localization.
The training procedures of \emph{GeoReasoner} are divided into two folds: reasoning tuning and location tuning.
In the first stage, we utilize the 3K reasoned text-image pairs encapsulating human inference knowledge, to fine-tune a well-trained LVLM model with LoRA~\cite{hu2021lora} for reasoning adaptation.
In the second stage, we leverage the curated 70K high-locatability GSV images dataset, to further fine-tune the LVLM model with another LoRA stacked on the first one for location tuning.
We assess \emph{GeoReasoner} in terms of accuracy for both country-level (\ie, predicting the country in which a street view is located) and city-level (\ie, predicting the city in which a street view is located) geo-localization. 
The results demonstrate that \emph{GeoReasoner} outperforms the other counterparts by more than \emph{25\%} at the country-level geo-localization and \emph{38\%} at the city-level geo-localization with reasoning on our test dataset.
Notably, \emph{GeoReasoner} performs slightly better than StreetCILP~\cite{haas2023learning}, which was trained on a substantially larger dataset of 1.1 million geo-tagged street-view images.
We also evaluate \emph{GeoReasoner} against state-of-the-art models for geo-localization using open benchmark datasets. The results show that \emph{GeoReasoner} achieves comparable performance with only 10k Flickr images used for training.
The main contributions of our work are:
\begin{itemize}
\vspace{-3mm}
\item We present a new paradigm that leverages an LVLM and external knowledge of human inference for geo-localization with reasoning from street-view images.
\item We introduce the concept of \emph{locatability} and devise a CLIP-based network to quantify the degree of locatability in street-view images.
\item We propose \emph{GeoReasoner}, an LVLM that outperforms existing geo-localization models and provides detailed reasoning for the inferred results. 
\end{itemize}
\section{Related work}
\subsection{Street Views}
Street views, as the realm of physical environments routinely accessed and engaged with in daily life, bear significant relevance to human perception~\cite{ye_2019_visual} and urban design~\cite{shen_2017_streetvizor}. 
Analyses of street views contribute to decision-making support~\cite{ye2019}, improved understanding of urban social and economic structures~\cite{bai2023}, and traffic asset monitoring and maintenance~\cite{campbell2019, li2021caption}.
This study places an emphasis on geo-localization based on street views.
Specifically, drawing motivation from~\citet{zhang2018}, we delineate the distribution of scene elements to quantify the degree of \emph{locatability} in street views.
Highly locatable street-view images are curated to train an LVLM that surpasses existing geo-localization models.

\subsection{Image-based Geo-localization}
Geo-localization entails determining spatial coordinates on the Earth's surface, with broad applications in practical scenarios, including tracking individual trajectories~\cite{cheng2022} and positioning autonomous vehicles~\cite{chalvatzaras2022}.
This study focuses on image-based geo-localization, utilizing image data as input.
Research on image-based geo-localization can be primarily classified into two approaches: retrieval-based~\cite{zhu2022, lin2022, zhang2023} and classification-based~\cite{clark2023, pramanick2022, muller2018, seo2018, weyand2016}.

The retrieval-based approach involves the sequential matching of a single image with a gallery of overhead views, each labeled with geographical coordinates, and identifying the result with the highest matching as the location. 
However, the utilization of this method is limited due to its requirement for additional reference datasets. 
The classification-based approach, exemplified by~\citet{weyand2016}, involves subdividing the Earth's surface into thousands of geographical cells and predicting the geographical unit to which an image belongs.
The prediction effectiveness can be boosted with a dataset comprising millions of street views, whilst the granularity is influenced by the number of subdivided geographical cells.
As such, many studies have been devoted to learning to corresponding multi-level features at different granularity~\cite{vo2017}, or multi-pair features for different tasks~\cite{muller2018,pramanick2022,cepeda2023}.

We approach image-based geo-localization with a novel paradigm.
Specifically, we integrate semantic visual concepts that offer locatable features~\cite{luo2022g3, theiner2022}, and incorporate human reasoning knowledge learned from geo-localization games using an LVLM.

\subsection{Vision-Language Models}
The emergence of Large Language Models (LLMs) has significantly impacted various tasks related to natural language processing.
These models exhibit remarkable performance in tasks such as text generation~\cite{zhang2023text} and text-based question answering~\cite{shao2023}, owing to their robust and versatile capabilities. 
As a result, research attention has shifted towards exploring prompt engineering techniques to enhance the performance of LLMs in downstream tasks~\cite{wei2022, yao2023tree, dai2023can, xu2023lvlm, ying2024mmt}. 

Large vision-language models (LVLMs) integrate visual encoders with LLMs, exhibiting remarkable effectiveness in visual question-answering tasks~\cite{liu2023llava, bai2023qwen, rao2023retrieval}.
This study harnesses the capabilities of LVLMs to address geo-localization of street views.
However, the optimal utilization of LVLMs remains a challenging issue, particularly due to the absence of high-quality training data and a lack of reasoning capabilities.
We overcome these challenges through an innovative paradigm and the thoughtful design of model architecture, contributing to a more effective utilization of LVLMs in this domain.

\section{GeoReasoner}
This section outlines our approach to addressing two challenges:
1) the absence of a high-quality street-view geo-localization dataset (discussed in Sect.~\ref{ssec:local}), and 2) the difficulty of integrating reasoning in geo-localization (discussed in Sect.~\ref{ssec:reason}), when constructing \emph{GeoReasoner}.

\begin{figure}[t]
  \centering
  \centerline{\includegraphics[width=0.99\columnwidth]{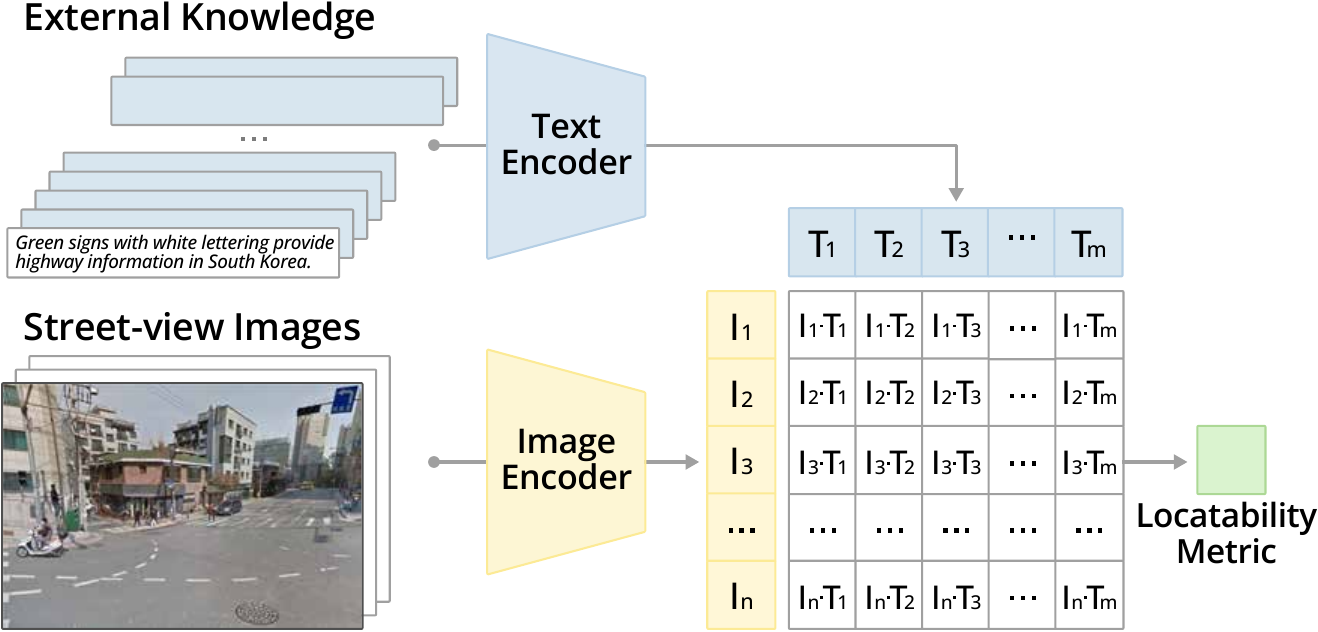}}
  \vspace{-2mm}
  \caption{The locatability quantization network devises a CLIP-based visual-text pairing approach to predict the locatability metric.}
  \vspace{-4mm}
  \label{fig:locatability}
\end{figure}

\begin{figure*}[t]
  \centering
  \includegraphics[width=0.99\textwidth]{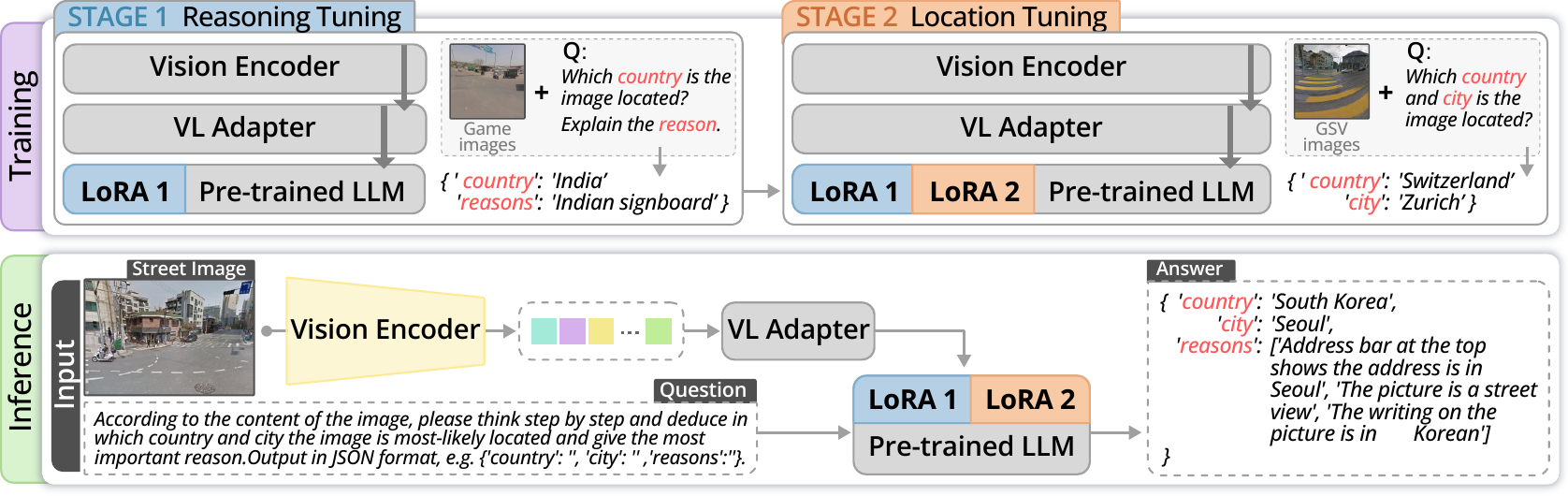}
  \vspace{-3mm}
  \caption{The architecture of \emph{GeoReasoner} consists of three modules: \emph{Vision Encoder}, \emph{VL Adapter} and \emph{Pre-trained LLM}. The model undergoes a two-fold supervised fine-tuning process: reasoning tuning and location tuning, to enable geo-localization with reasoning. }
  \vspace{-3mm}
  \label{fig:model_arch}
\end{figure*}

\subsection{Locatability-Enhanced Data Curation}
\label{ssec:local}
\begin{figure}[t]
  \begin{center}
  \includegraphics[width=0.95\columnwidth]{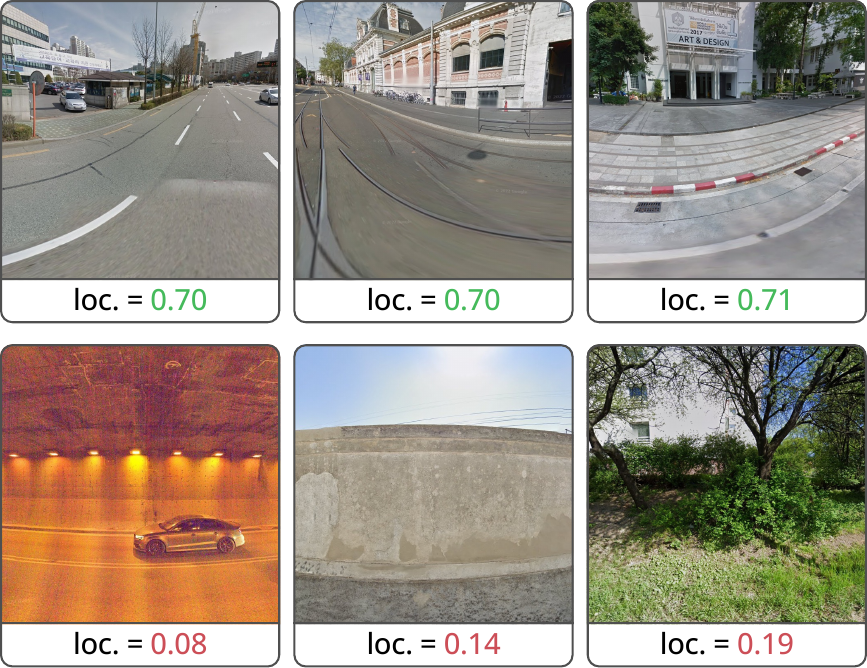}
  \vspace{-2mm}
  \caption{Locatability examples. Top row: the street views are highly locatable by signboards, architectural styles, and landmarks.
  Bottom row: no visual clues for locating the street views.}
  \vspace{-6mm}
  \label{fig:loc}
  \end{center}
\end{figure}

Throughout the development of this work, we observed variations in the degree of locatability among different street views.
For example, the images featuring textual signboards or prominent landmarks (\eg, Eiffel Tower) are easily locatable, whilst those captured in a tunnel or obscured by a wall tend to be less locatable. 
Refer to Figure~\ref{fig:loc} for further illustration.
Simply merging all these street-view images to train an LVLM is not an optimal approach, as the inclusion of poor-quality data can adversely affect the training efficiency of updating an LVLM~\cite{radford2021learning}.
To this end, we introduce \emph{locatability}, a metric that quantifies the level of locatability of street-view images.
We then devise a CLIP-based visual-text pairing network to produce the desired locatability metric for an input street-view image, as shown in Figure~\ref{fig:locatability}.
The network naturally incorporates data from two perspectives:

\begin{itemize}
\vspace{-3mm}
\item
\textbf{Street-View Images.}
We collected street-view images from the Google Street View\footnote{https://www.google.com/streetview} (GSV).
To enrich the diversity of the dataset, we first selected the top global cities according to the Globalization and World Cities Study Group and Network (GaWC) ranking.
Next, we utilized the global OpenStreetMap\footnote{https://www.openstreetmap.org} (OSM) geographic database to obtain the vector data of the road network in these urban areas.
The road network was passed to ArcPy, a Python site package of ArcGIS, to automatically extract sampling points at 4000-meter intervals and generate a CSV table containing information about these sampling points.
Subsequently, we employed the GSV API interface to compile a comprehensive dataset. This dataset encompassed street-view images captured from four distinct directions - front, back, left, and right - of each sampling point.
Considering the impact of data sparsity and image similarity, we randomly selected two of the four views from each data point, denoted as [$\textbf{I}_x, \textbf{I}_y$], where $x \in (left, right)$, $y \in (front, back)$. 
The process has yielded a total of over 130k street-view images with geo-tags collected from 72 cities in 48 countries.

\item
\textbf{Textual Clues.}
Textual clues often serve a pivotal role in delineating the geographical locations of street-view images. 
Two prominent games, GeoGuessr\footnote{https://www.geoguessr.com} and Tuxun\footnote{https://tuxun.fun}, which focus on geo-localization through street views, offer a potential solution to this gap. 
Their communities have collaboratively curated a well-organized collection of textual clues, used for pinpointing geographical locations across various countries and cities.
These clues, maintained by both players and administrators, provide valuable domain knowledge that aids in identifying and evaluating key geographical features in street views.
While such datasets now exist~\cite{luo2022g3}, there are no readily available image-text data pairs specifically tailored for LVLM training.
To bridge this gap, we gathered image-text pairs for geo-localization from these two open-source communities.
Subsequently, we utilized the BERT-based Named Entity Recognition (NER)~\cite{kenton2019} model to clean and filter text that lacked specific geographical location information. 
In this way, we collected a total of over 3K textual clues that encapsulate rich geo-localization information.
For instance, \emph{``houses in central Chile are more likely to have terracotta tiled roofs"}.
Each clue is paired with a corresponding street-view image.

\end{itemize}

With the GSV images and textual clues, our subsequent task is to filter GSV images with a high degree of locatability, for the purpose of training an LVLM.
To achieve this, we design a CLIP-based visual-text pairing network.
As depicted in Figure~\ref{fig:locatability}, the GSV images undergo processing by an image encoder that deduces the image attributes.

Here, we first use MaskFormer~\cite{cheng2021} to predict segmentation masks for various classes in GSV images, such as buildings, sky, and vehicles. We then compute an $n$-length vector $\textbf{I}_{seg}$, which quantifies the area ratio of each mask class, where $n$ represents the number of classes.
Subsequently, we utilize Sentence-BERT~\cite{reimers2019} to measure the similarity between textual clues and semantic segmentation labels, yielding an $m \times n$ matrix $M$, where $m$ is the number of textual clues.
After that, we normalize $M$ using min-max normalization, and set values lower than the threshold to zero, resulting in another \( m \times n \) matrix $\hat{M}$. We reduce \(\hat{M}\) to an \( n \)-length vector by calculating the mean across its rows, and then normalize it to obtain \(\textbf{w}_{loc}\). 
This vector represents the importance of each semantic segmentation label for geo-localization.
With the segmentation mask ratio $\textbf{I}_{seg}$ and the corresponding weight $\textbf{w}_{loc}$, the locatability metric of a GSV image is computed through the multiplication and accumulation of the respective values, as follows:
\begin{equation}
locatability(\textbf{I}_{seg}, \textbf{w}_{loc}) = \sum_{k=1}^n \textbf{I}_{seg}(k) \cdot \textbf{w}^k_{loc}, 
\label{eq:local}
\vspace{-3mm}
\end{equation}

where $\textbf{I}_{seg}(k)$ denotes pixel ratio of the $k$-th class in the segmentation mask $\textbf{I}_{seg}$.

A higher $locatability$ value indicates a higher degree of visual clues exhibited in a GSV image for geo-localization, while a lower value suggests the opposite.
Empirically, we selected a threshold value of 0.4 for filtering locatable GSV images.
This resulted in over 70k highly locatable images with geo-tags passing to the next stage for training an LVLM.

\subsection{Geo-localization with Reasoning}
\label{ssec:reason}

While many models (\eg,~\citet{clark2023, pramanick2022, muller2018, seo2018, weyand2016}) exist for image-based geo-localization, these models typically predict locations without providing the inference process.
This introduces several limitations:
First, the models operate as black boxes without providing insights, making it challenging for users to interpret.
This obstacle impedes further refinement of the geo-localization model.
More importantly, studies have demonstrated that integrating the reasoning process can enhance the capabilities of LLMs~\cite{qiao2022reasoning}.
Therefore, our objective is to construct an LVLM for image-based geo-localization with reasoning capability.

\textbf{Model Architecture.}
Figure~\ref{fig:model_arch} illustrates the architecture of the proposed model \emph{GeoReasoner}, which is based on Qwen-VL~\cite{bai2023qwen}.
\emph{GeoReasoner} consists of three modules: \emph{Vision Encoder}, \emph{Vision-Language (VL) Adapter} and \emph{Pre-trained LLM}.
Specifically, the \emph{Vision Encoder} module employs the Vision Transformer (ViT)~\cite{dosovitskiy2020image} architecture.
The input street-view images are resized to a specific resolution and then divided into a set of image patches.
To refine image patches into sequential representations compatible with an LLM, the \emph{VL Adapter} is introduced.
In the VL Adapter, the sequence of visual features is initially condensed to a fixed length to address efficiency challenges posed by the substantial number of visual feature sequences.
Subsequently, the processed visual features are integrated with the LLM using cross-attention mechanisms.
Following this, the compressed visual feature sequence and text sequence are passed to the \emph{Pre-trained LLM} module, which functions as a decoder for generating the answer.

\textbf{Supervised Fine-tuning.}
The overall model undergoes a staged pre-training process that is divided into two folds: reasoning tuning and location tuning.
In the first stage, our objective is to enhance the model's reasoning capability by utilizing textual clues paired with street-view images collected from geo-localization games.
The input street-view image \& question, and the output answer are formatted as prompts in the following manner:

  \vspace{2mm}
  \centerline{\includegraphics[width=0.9\columnwidth]{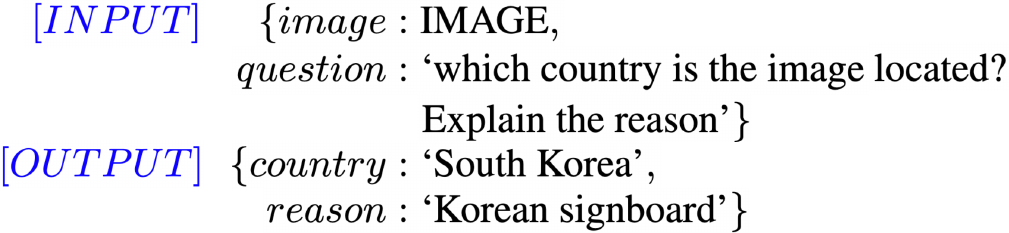}}
  \vspace{-2mm}

Here, we can only provide reasoning at the country level due to the granularity exhibited in the image-text pairs.
Nevertheless, this reasoning procedure is sufficient to facilitate the second stage of location tuning.
Next, we integrate the prior knowledge of country information with highly locatable GSV images with geo-tags to infer the fine-grained city-level location information.
We utilize a similar prompt format as in the first stage but without a reasoning requirement.
Both stages are fine-tuned from the pre-trained Qwen-VL with LoRA, which contributes to the overall performance improvement of Qwen-VL in both the reasoning and location tuning stages, allowing the model to better capture complex relationships within the image-text pairs.

\section{Experiments}
We conduct a series of experiments to evaluate the effectiveness of the locatability-enhanced geo-localization dataset (Sect.~\ref{ssec:loc_exp}) and the model \emph{GeoReasoner} for geo-localization with reasoning (Sect.~\ref{ssec:reason_exp}).

\subsection{Experiments on Locatability-Enhanced Dataset} \label{ssec:loc_exp}

\begin{figure}[t]
  \begin{center}
  \includegraphics[width=0.88\columnwidth]{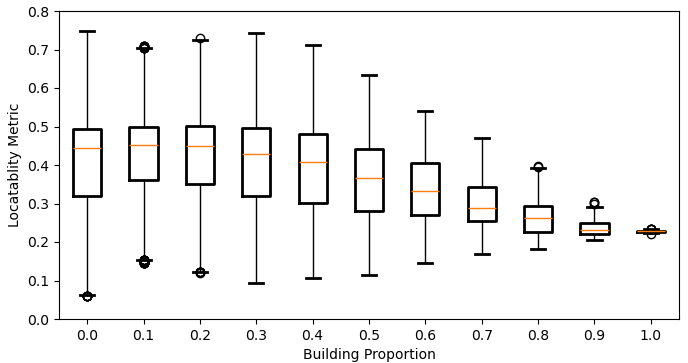}
  \vspace{-3mm}
  \caption{The relationship between building proportion and the degree of locatability in street views. The locatability metric peaks when the building proportion is approximately 0.2.}
  \vspace{-5mm}
  \label{fig:building}
  \end{center}
\end{figure}

\subsubsection{Qualitative Comparison}
Figure~\ref{fig:loc} presents examples of the predicted locatability degrees of different street-view images by our locatability quantization network.
The top row showcases street views distinguished by prominent localizable attributes.
The left image features the Korean language on a signboard, the middle image captures the distinctive \emph{Art Nouveau} architectural style commonly found in Switzerland, and the right image shows an art \& design museum in India.
In contrast, street views in the bottom row display lower locatability degrees.
The left image resembles a tunnel, lacking additional discernible information for accurate localization.
Similarly, the middle image is occluded by a wall, and the right image faces common vegetation that is available worldwide.

\begin{figure}[b]
  \vspace{-3mm}
  \begin{center}
  \includegraphics[width=0.95\columnwidth]{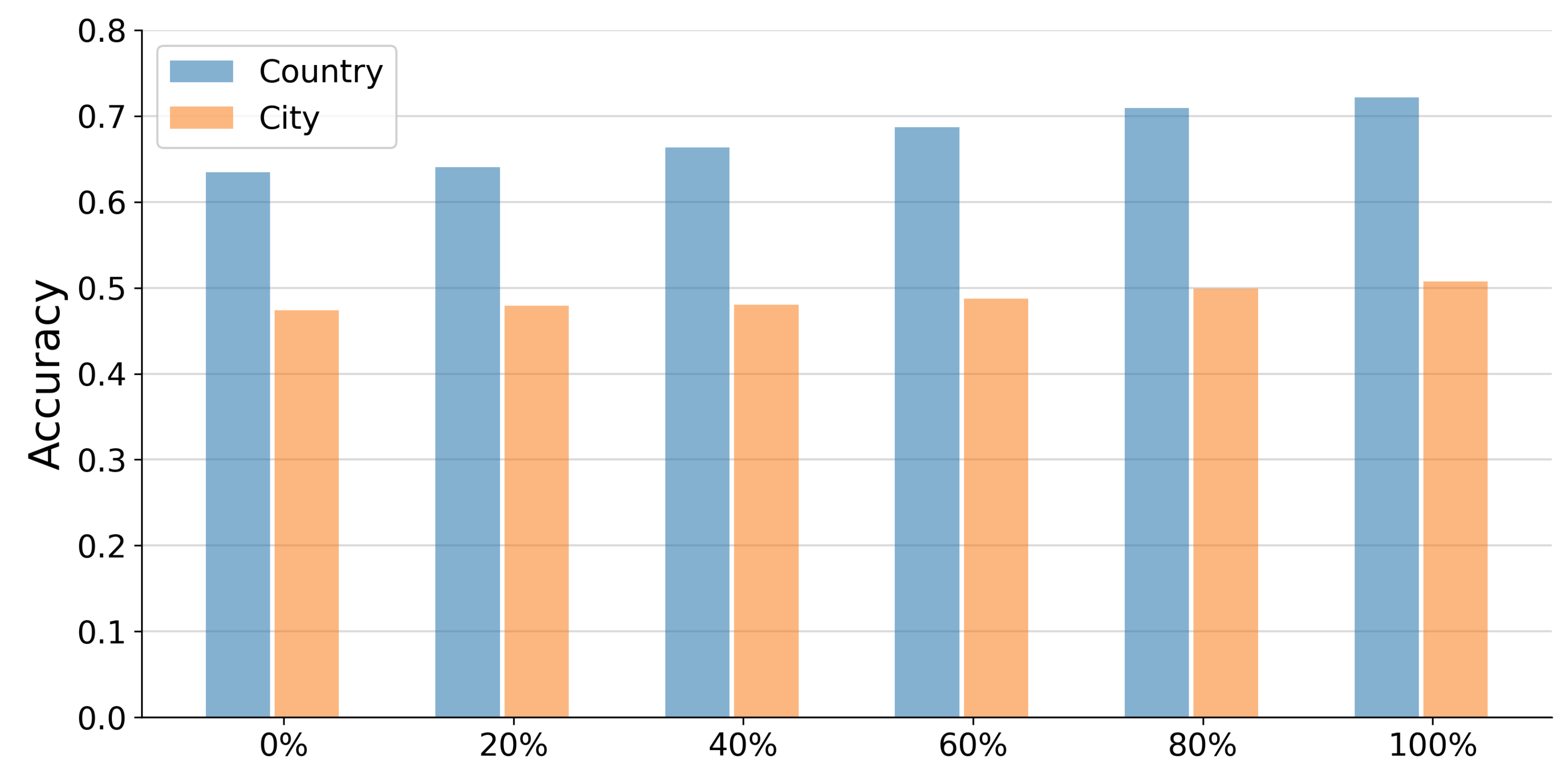}
  \vspace{-2mm}
  \caption{Quantitative comparison of country- and city-level geo-localization accuracy by different models trained on mixed datasets with varying proportions of high locatable GSV images.}
  \label{fig:vis}
  \end{center}
\end{figure}

\begin{figure*}[t]
  \begin{center}
  \includegraphics[width=0.98\textwidth]{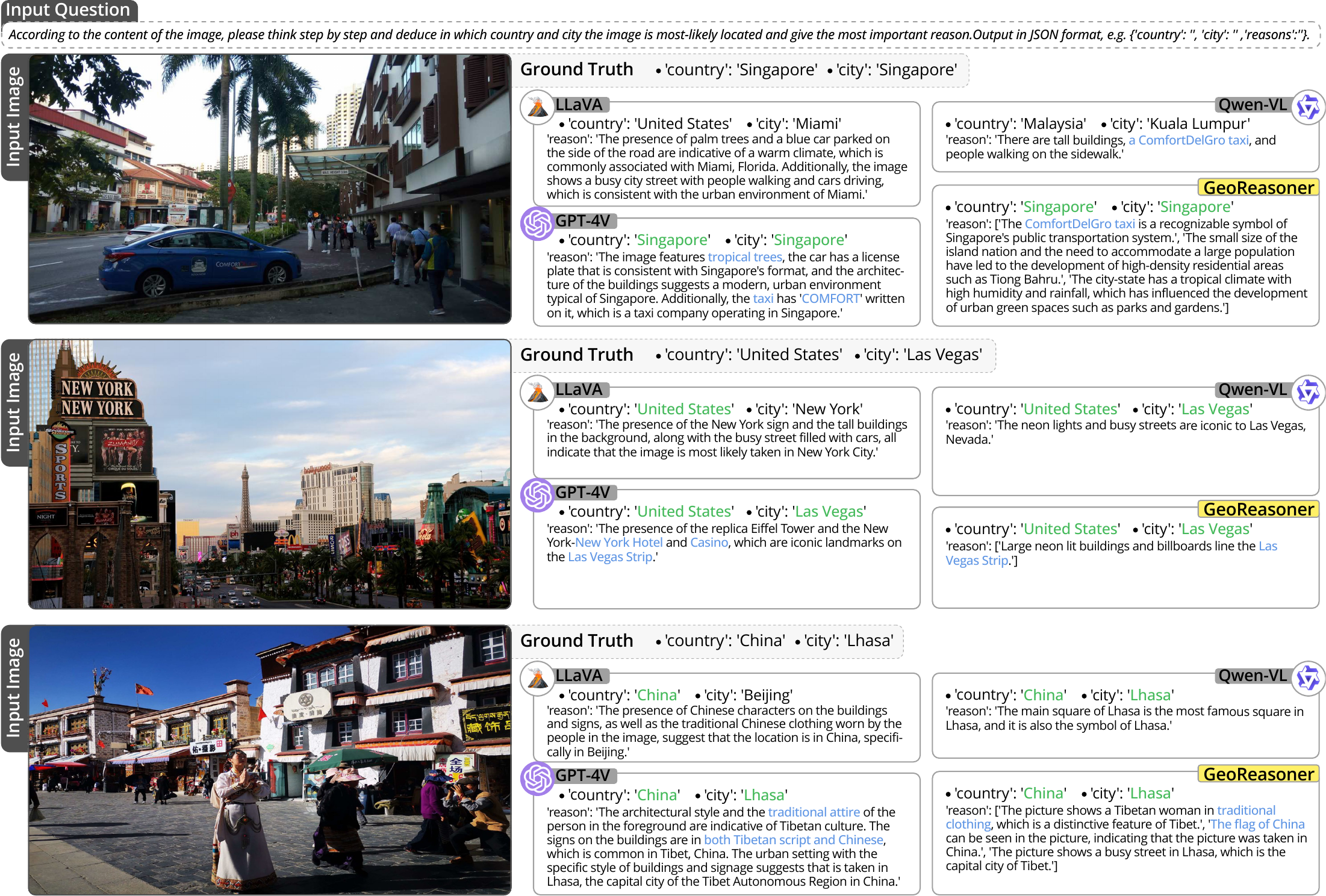}
  \vspace{-3mm}
  \caption{Examples of LVLM-based approaches in geo-localization with reasoning. Prediction results matching the ground truth are highlighted in \textcolor{green}{green}, while reasons offering valid information are marked in \textcolor{blue}{blue}.}
  \vspace{-4mm}
  \label{fig:reason_comp}
  \end{center}
\end{figure*}

For the proposed locatability metric in Equation~\ref{eq:local}, we also evaluated the relationship between building proportion and the degree of locatability of street views. 
The results are shown in Figure~\ref{fig:building}. 
The locatability metric slightly increases as the building proportion ranges from 0 to 0.2, but decreases as the building proportion continues to increase.
The results indicate that buildings are not the sole determinant of locatability.  
As the proportion of buildings increases, the street-view images transition from panoramic to close-up views, leading to reduced information availability and consequently diminishing the degree of locatability.

The qualitative analysis indicates the effectiveness of the locatability quantization network in predicting locatability degrees of street-view images.
Furthermore, the prediction aligns with human inference knowledge harvested from real geo-localization games, providing the ground truths for fine-tuning the reasoning component in \emph{GeoReasoner}.

\subsubsection{Quantitative Comparison}\label{sssec:data_exp}

We conducted quantitative experiments to investigate the importance of using high-locatability GSV images in training the location component in \emph{GeoReasoner}.
Various datasets were prepared, featuring different proportions of high-locatability GSV images, ranging from 0\% (only low-locatability GSV images) to 100\% (only high-locatability GSV images).
To ensure fairness, each experimental group retained consistent 10K GSV images, with only the proportion of high-locatability images varying. 
Subsequently, models were trained for each dataset, and their accuracy in country- and city-level geo-localization was evaluated on a randomly sampled set of 1K GSV images.

The experimental results are presented in Figure~\ref{fig:vis}.
Overall, the results reveal that as the proportion of high-locatability GSV images in the training dataset increases, the performance of the fine-tuned location component improves in both country- and city-level geo-localization.
Specifically, the country- and city-level geo-localization accuracy increases from 0.63 \& 0.47 for 0\% high-locatability GSV images, to 0.72 \& 0.51 for 100\% high-locatability GSV images.
Notably, the experiments only utilize 10K GSV images instead of all the curated 70K high-locatability GSV images due to training complexity.
Nevertheless, the results demonstrate that high-locatability GSV images offer more meaningful insights and less extraneous noise, making them highly valuable in the geo-localization task.

\begin{table*}
\footnotesize
\vspace{-3mm}
\caption{Comparison of Precision, Recall and F1 scores in country-level and city-level geo-localization. * represents the model trained on high-locatability GSV images.} 
\label{tab:comp} 
\begin{center} 
\begin{tabular}{lcccccc} 
\toprule 
\multirow{2}*{Model} & \multicolumn{3}{c}{Country} & \multicolumn{3}{c}{City} \\ 
& Precision$\uparrow$ & Recall$\uparrow$ & F1$\uparrow$ & Precision$\uparrow$ & Recall$\uparrow$ & F1$\uparrow$ \\ 
\midrule 
StreetCLIP~\cite{haas2023learning} & 0.7943 & 1.00 & 0.8854 & 0.7457 & 1.00 & 0.8543\\
LLaVA~\cite{liu2023llava} & 0.4029 & 1.00 & 0.5744 & 0.2400 & 1.00 & 0.3871\\
Qwen-VL (Qwen-7B)~\cite{bai2023qwen} & 0.5829 & 0.95 & 0.7225 & 0.3743 & 0.89 & 0.5270\\
GPT-4V~\cite{gpt4v} & 0.8917 & 0.34 & 0.4923 & 0.5083 & 0.31 & 0.3851\\
ViT*~\cite{dosovitskiy2020image} & 0.7100 & 1.00 & 0.8304 & 0.6762 & 1.00 & 0.8068\\
GeoReasoner* & 0.8237 & 1.00 & \textbf{0.9033} & 0.7521 & 1.00 & \textbf{0.8585}\\
\bottomrule 
\end{tabular} 
\vspace{-4mm}
\end{center} 
\end{table*}

\subsection{Experiments on Geo-localization with Reasoning} \label{ssec:reason_exp}

\subsubsection{Qualitative Comparison with SOTA}
To assess the efficacy of \emph{GeoReasoner} in terms of geo-localization with reasoning, we conduct a qualitative comparison with state-of-the-art LVLM-based approaches, including LLaVA~\cite{liu2023llava}, Qwen-VL (Qwen-7B)~\cite{bai2023qwen}, and GPT-4V~\cite{gpt4v}.
In the experimental phase, we presented the same input street-view images, reasoning process, and result formats to these models.
Specifically, a consistent prompt is used, as below: 

\emph{According to the content of the image, please think step by step and deduce in which country and city the image is most likely located and offer possible explanations.
Output in JSON format, e.g., \{`country': `', `city': `', `reasons':`'\}}.

Figure~\ref{fig:reason_comp} illustrates the inference results of counterpart models and \emph{GeoReasoner} on three diverse street views from different countries and cities—namely, Singapore-Singapore (top), United States-Las Vegas (middle) and China-Lhasa (bottom).
Overall, \emph{GeoReasoner} not only outperforms existing models in the accuracy of country or city-level predictions but also provides coherent explanations with insightful reasoning for the inference results.

In Figure~\ref{fig:reason_comp} (top), \emph{GeoReasoner} identifies the word `COMFORT' on the taxi in the image.
Drawing from prior knowledge, \emph{`the ComfortDelGro taxi is a distinctive symbol of Singapore's public transportation system'} in the text-image pairs, the model deduces that the area is likely to be in \textit{Singapore}.
GPT-4V predicts the same geo-location with accurate reasoning, yet the other two models fail, either due to not recognizing the taxi by LLaVA or making an incorrect inference about the city by Qwen-VL.

Figure~\ref{fig:reason_comp} (middle) presents a scene of the Las Vegas Strip.
A conspicuous `NEW YORK' sign is prominently visible in the upper-left corner of the image. 
This sign causes the reasoning error in the task performed by LLaVA.
Although Qwen-VL generates accurate predictions of \emph{Las Vegas-United States}, the most essential factor, \ie, `Las Vegas Strip', is not considered in the reasoning process.
In contrast, both \emph{GeoReasoner} and GPT-4V provide the correct geo-location along with accurate inference.

Based on the depiction of Chinese characters and traditional clothing in Figure~\ref{fig:reason_comp} (right), all models make accurate predictions regarding the country, identifying it as \emph{China}.
However, LLaVA makes an incorrect prediction of the city, specifying \emph{Beijing}.
In contrast, the other models successfully predict the city as \emph{Lhasa}, providing sensible and justifiable reasons for their inferences.

\subsubsection{Quantitative Comparison with SOTA}\label{sssec:quan_sota}

We further conduct quantitative experiments to compare with counterparts LVLMs.
In addition, we choose StreetCLIP~\cite{haas2023learning} as the state-of-the-art classification-based approach and omit retrieval-based approaches relying on a geo-tagged image gallery that is not available.
It is important to clarify that, for the LVLM-based approaches, obtaining corresponding and relevant answers is not guaranteed at all times. 
Therefore, we included \emph{Recall} rate to measure the proportion of effective answers within the large language models.
When calculating the \emph{Accuracy} rate, only the accuracy of these effective answers is taken into account.
We additionally compute \emph{F1} values, taking into consideration both \emph{Accuracy} and \emph{Recall} metrics.

Table~\ref{tab:comp} presents the prediction results by the counterparts and \emph{GeoReasoner}.
Overall, \emph{GeoReasoner} outperforms all the counterparts, particularly those LVLM-based approaches.
Taking the best performed Qwen-VL for example, \emph{GeoReasoner} outperforms it 25.02\% on country-level geo-localization and 38.61\% on city-level geo-localization, in terms of F1 value. 
Surprisingly, the recall performance of GPT-4V for the geo-localization task was notably low.
Most of the responses were mainly: \emph{`I'm sorry, I can't provide assistance with that request.'} or \emph{`I'm sorry, but I am unable to provide the exact location, such as the country and city, for the image you have provided. My capabilities do not include analyzing specific details to determine the geographical location of the image content.'}
{\scriptsize
\begin{table*}[h]
\vspace{-3mm}
\caption{Results of the ablation experiments using baseline Qwen-VL (Qwen-7B), GeoReasoner w/o location tuning, GeoReasoner w/o reasoning tuning, and the full GeoReasoner models.}
\label{tab:ablation}
\begin{center}
\footnotesize
\begin{tabular}{lcccccccc} 
\toprule
\multirow{3}*{Model} & \multicolumn{2}{c}{Training} & \multicolumn{6}{c}{Performance}\\
& \multirow{2}*{Reasoning} & \multirow{2}*{Location}   & \multicolumn{3}{c}{Country} & \multicolumn{3}{c}{City}\\
&  &  & Precision$\uparrow$ & Recall$\uparrow$ & F1$\uparrow$ & Precision$\uparrow$ & Recall$\uparrow$ & F1$\uparrow$\\
\midrule
Qwen-VL (Qwen-7B) & - & - & 0.5829 & 0.95 & 0.7225 & 0.3743 & 0.89 & 0.5270\\
GeoReasoner w/o location tuning & \checkmark & $\times$  & 0.6971 & 1.00 & 0.8215 & 0.4114 & 0.99 & 0.5813\\
GeoReasoner w/o reasoning tuning & $\times$ & \checkmark & 0.7803 & 1.00 & 0.8766 & 0.7029 & 1.00 & 0.8255\\
GeoReasoner & \checkmark & \checkmark & \textbf{0.8237} & 1.00 & \textbf{0.9033} & \textbf{0.7521} & 1.00 & \textbf{0.8585}\\  
\bottomrule
\end{tabular}
\vspace{-6mm}
\end{center}
\end{table*}
}

We speculate that GPT-4V has undergone extensive measures to ensure the model's security and privacy, which may contribute to its reluctance or denial of recognition in the task of geo-localization.

In comparison to StreetCLIP that is specialized in geo-localization, \emph{GeoReasoner} demonstrates only a slight superiority.
Nevertheless, it's important to note that StreetCLIP was trained on a significantly larger dataset of over 1.1 million street-view images, while our \emph{GeoReasoner} was trained with only 70K street views.
For ViT trained on the same data, \emph{GeoReasoner} still exhibits superior geolocation capabilities.
Moreover, \emph{GeoReasoner} offers reasoning capability, providing added value for various downstream tasks.

{\scriptsize
\begin{table}[t]
\footnotesize
\caption{Comparison results on Im2GPS dataset. 
The top five rows are derived from the results reported in the paper, while the last four rows are from retesting on the filtered Im2GPS dataset, which includes only highly locatable data.} 
\vspace{-4mm}
\label{tab:im2gps} 
\begin{center} 
\begin{tabular}{lccccc} 
\toprule 
\multirow{2}*{Model} & \multicolumn{2}{c}{Dataset w/ Filter} & Street & City & Country \\ 
 & Train & Test & 1km & 25km & 750km \\ 
\midrule 
PlaNet & $\times$ & $\times$ & 0.08 & 0.25 & 0.54 \\
CPlaNet & $\times$ & $\times$ & 0.17 & 0.37 & 0.62 \\
ISNs & $\times$ & $\times$ & 0.17 & 0.43 & 0.67 \\
Translocator & $\times$ & $\times$ & 0.20 & 0.48 & 0.76 \\
GeoDecoder & $\times$ & $\times$ & 0.22 & 0.50 & 0.80 \\
\midrule
ISNs & $\times$ & \checkmark & 0.25 & 0.43 & 0.78 \\
GeoCLIP & $\times$ & \checkmark & 0.25 & 0.49 & 0.87 \\
GeoReasoner & $\times$ & \checkmark & 0.10 & 0.41 & 0.82 \\
GeoReasoner & \checkmark & \checkmark & 0.13 & 0.44 & 0.86 \\
\bottomrule 
\end{tabular} 
\vspace{-6mm}
\end{center} 
\end{table}
}

\subsubsection{Ablation Experiments}\label{sssec:abla}
To assess the contributions of the location tuning and reasoning tuning components in \emph{GeoReasoner}, we design several ablation experiments using the Qwen-VL~\cite{bai2023qwen} pre-trained model as the baseline.
Next, we integrated the Qwen-VL pre-trained model with LoRA1 (\emph{GeoReasoner} without location tuning) and LoRA2 (\emph{GeoReasoner} without reasoning tuning).  
The last experiment involved the full \emph{GeoReasoner} model, including both the location tuning and reasoning tuning components.
The same prompts were utilized for all these models, as in the previous experiments.

Table~\ref{tab:ablation} presents the quantitative results in terms of \emph{accuracy}, \emph{recall}, and \emph{F1}.
Overall, the results indicate that both the location tuning and reasoning tuning components improve the model performance.
Specifically, the location tuning component is essential for geo-localization, as \emph{GeoReasoner} w/o reasoning tuning (row 3) achieves much higher accuracy than \emph{GeoReasoner} w/o location tuning (row 2), especially for fine-grained city-level prediction.
This result further strengthens the evidence that high-locatability GSV images are essential for geo-localization.
The reasoning tuning component also plays a significant role in the performance improvement, as evidenced by the superior performance of the full \emph{GeoReasoner} (row 4).

{\scriptsize
\begin{table}[t]
\footnotesize
\caption{Comparison results on Im2GPS3k dataset. 
The top six rows are derived from the results reported in the paper, while the last four rows are from retesting on the filtered Im2GPS3k dataset, which
includes only highly locatable data.} 
\vspace{-2mm}
\label{tab:im2gps3k} 
\begin{center} 
\begin{tabular}{lccccc} 
\toprule 
\multirow{2}*{Model} & \multicolumn{2}{c}{Dataset w/ Filter} & Street & City & Country \\ 
 & Train & Test & 1km & 25km & 750km \\ 
\midrule 
PlaNet & $\times$ & $\times$ & 0.09 & 0.25 & 0.48 \\
CPlaNet & $\times$ & $\times$ & 0.10 & 0.27 & 0.49 \\
ISNs & $\times$ & $\times$ & 0.11 & 0.28 & 0.50 \\
Translocator & $\times$ & $\times$ & 0.12 & 0.31 & 0.59 \\
GeoDecoder & $\times$ & $\times$ & 0.13 & 0.34 & 0.61 \\
GeoCLIP & $\times$ & $\times$ & 0.14 & 0.34 & 0.70 \\
\midrule 
ISNs & $\times$ & \checkmark & 0.10 & 0.29 & 0.59 \\
GeoCLIP & $\times$ & \checkmark & 0.12 & 0.38 & 0.83 \\
GeoReasoner & $\times$ & \checkmark & 0.09 & 0.35 & 0.74 \\
GeoReasoner & \checkmark & \checkmark & 0.10 & 0.38 & 0.83 \\
\bottomrule 
\end{tabular} 
\vspace{-6mm}
\end{center} 
\end{table}
}

\subsubsection{Generalizability Evaluation}
To further assess the generalizability of \emph{Georeasoner} in geo-localization, we conduct additional testing on open Flickr image datasets of Im2GPS~\cite{hays2008im2gps} and Im2GPS3k~\cite{vo2017}.
Here we use only 10k Flickr images for fine-tuning \emph{Georeasoner}.
Since \emph{Georeasoner} predicts city names rather than GPS coordinates, we first convert the predicted city names generated by \emph{Georeasoner} into the GPS coordinates of their respective city centers, then measure the distance between these predicted coordinates and ground-truth locations.

Table~\ref{tab:im2gps} and Table~\ref{tab:im2gps3k} present the performance comparison of \emph{Georeasoner} with PlaNet~\cite{weyand2016}, CPlaNet~\cite{seo2018}, ISNs~\cite{muller2018}, Translocator~\cite{pramanick2022}, GeoDecoder~\cite{clark2023}, and GeoCLIP~\cite{cepeda2023} on Im2GPS and Im2GPS3k datasets, respectively.
The results demonstrate that fine-tuning \emph{GeoReasoner} using highly locatable images significantly improves prediction accuracy for street, city, and country levels (row 8 vs. row 9 in Table~\ref{tab:im2gps}, and row 9 vs. row 10 in Table~\ref{tab:im2gps3k}).
Remarkably, despite being fine-tuned solely on a smaller number of Flickr images, \emph{GeoReasoner} achieves results comparable to ISNs and GeoCLIP trained on millions of Flickr images, particularly in terms of city- and country-level accuracy.
Besides, \emph{GeoReasoner} trained on the filtered, highly locatable Flickr images also show improvements in the city- and country-level geo-localization, demonstrating the generalizability of our proposed \emph{locatability} module.
\section{Discussion}
\textbf{The significance of high-locatability street-view images.}
We observe a significant performance improvement when \emph{GeoReasoner} is trained upon high-locatability street-view images. 
Such images often contain explicit visual clues such as stylized architecture, traffic signs, and landmarks, providing the model with richer contextual information.
Therefore, increasing the quality of the training dataset enhances the model's geo-localization performance.
Additionally, the quantity of high-locatability images is vital, as the model trained with 70K images (as in Sect.~\ref{sssec:quan_sota}) achieves significantly higher accuracy than the one trained with 10K images (Sect.~\ref{sssec:data_exp}).
In balancing the quality and quantity of the training dataset, we empirically applied a threshold of 0.4 to differentiate between highly and less localizable street views.
Setting the threshold too high (\eg, 0.7) can lead to a notable decrease in the amount of high-locatability images, whilst a lower threshold (\eg, 0.1) may bring in introduce low-quality images.

\textbf{The necessity of reasoning process.}
The introduction of the reasoning component successfully elevated \emph{GeoReasoner}'s performance in the geo-localization task.
This signifies that LVLM can adeptly capture intricate relationships among image features, location clues, and geo-locations in the training process.
Implemented an innovative solution to empower the reasoning capability within \emph{GeoReasoner} by leveraging human inference knowledge extracted from geo-localization games.
Despite the relatively small dataset, a noticeable improvement in performance has been achieved.
In the future, we plan to expand the reasoning dataset by diversifying the influencing clues.
For instance, the current textual clues are absent of landscape information, which could provide invaluable insights for geo-localization.
We will collaborate with domain experts such as urban planners and geographers to address these limitations.

\begin{figure}[t]
  \begin{center}
  \includegraphics[width=0.99\columnwidth]{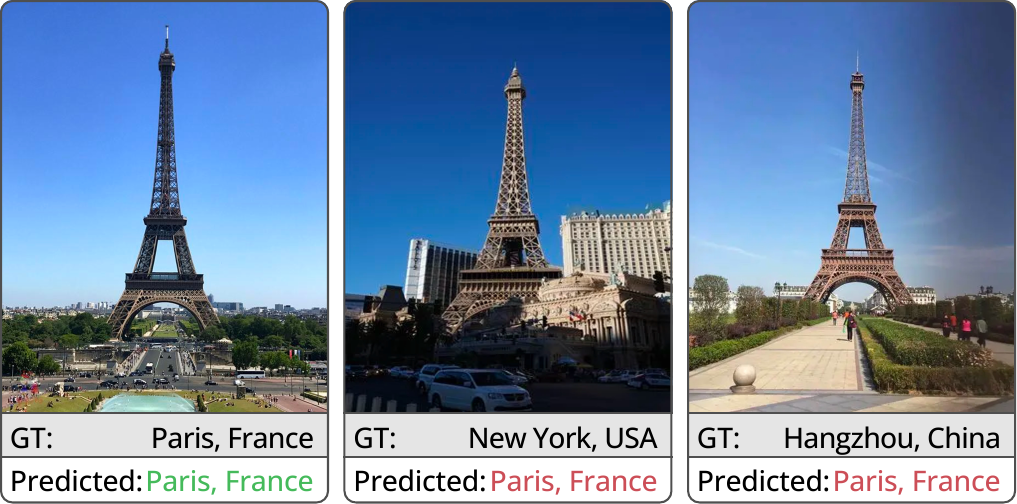}
  \vspace{-2mm}
  \caption{GeoReasoner fails to distinguish the Eiffel Tower and its replicas in New York, USA, and Hangzhou, China.}
  \vspace{-5mm}
  \label{fig:7}
  \end{center}
\end{figure}

\textbf{Failure cases.}
\emph{GeoReasoner} comprehends architectural style as a pivotal factor in geo-localization.
However, the model can be misled by the learned significance of architectural style.
Figure~\ref{fig:7} presents a street view of the Eiffel Tower in Paris, France (left), and replicas of the Eiffel Tower in New York, USA (middle) and in Hangzhou, China (right).
\emph{GeoReasoner} fails to distinguish between them, predicting all instances as located in Paris, France.
This misclassification is not unique to \emph{GeoReasoner} but also extends to other LVLMs like GPT-4V. Consequently, it underscores the necessity for LVLM-based methods to delve deeper into knowledge for more sophisticated geo-localization capabilities.
Once again, it is imperative to collaborate with domain experts and enhance the visual clues and reasoning procedure comprehensively to tackle this issue.

\section{Conclusion}
In this paper, we present a new paradigm that integrates a large vision-language (LVLM) model with human inference knowledge for street view geo-localization with reasoning.
We introduce the concept of \emph{locatability} and devise a CLIP-based network to quantify the degree of locatability in street-view images, facilitating the selection of high-quality data.
We design an LVLM-based model named \emph{GeoReasoner}, which harnesses external knowledge of human inference from real geo-localization games and curated high-quality data to enhance the performance of geo-localization tasks with reasoning capabilities.
The model undergoes two-stage fine-tuning, namely \emph{reasoning tuning} and \emph{location tuning}.
The \emph{reasoning tuning} stage aims to acquire potential linkage between coarse-grained geographical locations (\ie, country) and the associated positioning reasons.
In \emph{location tuning} stage, we employ the curated high-quality data to further refine the model in fine-grained geo-localization (\ie, city) learning.
Extensive experiments prove that \emph{GeoReasoner} outperforms previous models qualitatively and quantitatively.

\section*{Acknowledgements}
We would like to thank Yao Zhou and Wenqi Shao for their insightful discussions and Ziyao Gao for her assistance in drawing the figures in this paper. 
We also extend our gratitude to the anonymous reviewers for their valuable comments.
This work is partially supported by the National Natural Science Foundation of China (62172398, 42171456, 52078343).

\section*{Impact Statement}
\emph{GeoReasoner} advances image-based geo-localization technologies that are pivotal for many applications such as autonomous navigation.
The pipeline of constructing the dataset featuring high-locatability street views proves highly beneficial across multiple scenarios, such as urban studies, culture studies, and digital humanities, all of which are increasingly reliant on the analysis of high-quality street-view data.

The proposed paradigm represents the fusion of LVLM with human inference knowledge, which has implications for the advancement of artificial intelligence (AI) that is more aligned with human cognition. 
The synergy can lead to the creation of AI that is not only more effective in complex inference tasks but also more understandable and relatable to human users. 
As AI becomes more pervasive in daily life, the importance of designing systems that are both transparent and capable of complex reasoning cannot be overstated.

\bibliography{ref}

@String(TVCG = {IEEE Transactions on Visualization and Computer Graphics})

@String(ICML = {Proceedings of International Conference on Machine Learning})

@String(ICLR = {Proceedings of International Conference on Learning Representations})

@String(ACL = {Proceedings of the Annual Meeting of the Association for Computational Linguistics})

@String(NIPS = {Advances in Neural Information Processing Systems})

@String(CVPR = {Proceedings of the IEEE/CVF Conference on Computer Vision and Pattern Recognition})

@String(ECCV = {Proceedings of the European Conference on Computer Vision})

@String(ICCV = {Proceedings of the IEEE International Conference on Computer Vision})

@article{ye_2019_visual,
   author = {Ye, Yu and Zeng, Wei and Shen, Qiaomu and Zhang, Xiaohu and Lu, Yi},
   title = {The visual quality of streets: A human-centred continuous measurement based on machine learning algorithms and street view images},
   journal = {Environment and Planning B: Urban Analytics and City Science},
   volume = {46},
   number = {8},  
   year = {2019},
   pages = {1439-1457}
}

@article{shen_2017_streetvizor,
   author = {Shen, Qiaomu and Zeng, Wei and Ye, Yu and Mueller Arisona, Stefan and Schubiger, Simon and Burkhard, Remo and Qu, Huamin},
   title = {{StreetVizor: Visual Exploration of Human-Scale Urban Forms Based on Street Views}},
   journal = TVCG,
   volume = {24},
   number = {1},
   year = {2018},
   pages = {1004-1013}
}

@inproceedings{radford2021learning,
  title={{Learning Transferable Visual Models From Natural Language Supervision}},
  author={Radford, Alec and Kim, Jong Wook and Hallacy, Chris and Ramesh, Aditya and Goh, Gabriel and Agarwal, Sandhini and Sastry, Girish and Askell, Amanda and Mishkin, Pamela and Clark, Jack and others},
  booktitle=ICML,
  pages={8748-8763},
  year={2021},
  organization={PMLR}
}

@inproceedings{qiao2022reasoning,
  title={{Reasoning with Language Model Prompting: A Survey}},
  author={Qiao, Shuofei and Ou, Yixin and Zhang, Ningyu and Chen, Xiang and Yao, Yunzhi and Deng, Shumin and Tan, Chuanqi and Huang, Fei and Chen, Huajun},
  booktitle=ACL,
  pages={5368–5393},
  year={2023}
}

@article{zhang2018,
  title={Representing place locales using scene elements},
  author={Zhang, Fan and Zhang, Ding and Liu, Yu and Lin, Hui},
  journal={Computers, Environment and Urban Systems},
  volume={71},
  pages={153-164},
  year={2018},
  publisher={Elsevier}
}

@article{ye2019,
  title={Measuring daily accessed street greenery: A human-scale approach for informing better urban planning practices},
  author={Ye, Yu and Richards, Daniel and Lu, Yi and Song, Xiaoping and Zhuang, Yu and Zeng, Wei and Zhong, Teng},
  journal={Landscape and Urban Planning},
  volume={191},
  pages={103434},
  year={2019},
  publisher={Elsevier}
}

@article{bai2023,
  title={{Transport Object Detection in Street View Imagery Using Decomposed Convolutional Neural Networks}},
  author={Bai, Yunpeng and Shang, Changjing and Li, Ying and Shen, Liang and Jin, Shangzhu and Shen, Qiang},
  journal={Mathematics},
  volume={11},
  number={18},
  pages={3839},
  year={2023},
  publisher={MDPI}
}

@article{campbell2019,
  title={{Detecting and mapping traffic signs from Google Street View images using deep learning and GIS}},
  author={Campbell, Andrew and Both, Alan and Sun, Qian Chayn},
  journal={Computers, Environment and Urban Systems},
  volume={77},
  pages={101350},
  year={2019},
  publisher={Elsevier}
}

@article{cheng2022,
  title={{OPTDP: Towards optimal personalized trajectory differential privacy for trajectory data publishing}},
  author={Cheng, Wenqing and Wen, Ruxue and Huang, Haojun and Miao, Wang and Wang, Chen},
  journal={Neurocomputing},
  volume={472},
  pages={201-211},
  year={2022},
  publisher={Elsevier}
}

@article{chalvatzaras2022,
  title={{A Survey on Map-Based Localization Techniques for Autonomous Vehicles}},
  author={Chalvatzaras, Athanasios and Pratikakis, Ioannis and Amanatiadis, Angelos A},
  journal={IEEE Transactions on Intelligent Vehicles},
  volume={8},
  number={2},
  pages={1574-1596},
  year={2022},
  publisher={IEEE}
}

@article{cepeda2023,
  title={{GeoCLIP: Clip-Inspired Alignment between Locations and Images for Effective Worldwide Geo-localization}},
  author={Vivanco Cepeda, Vicente and Nayak, Gaurav Kumar and Shah, Mubarak},
  journal={Advances in Neural Information Processing Systems},
  volume={36},
  pages={},
  year={2024}
}

@article{cheng2021,
  title={{Per-Pixel Classification is Not All You Need for Semantic Segmentation}},
  author={Cheng, Bowen and Schwing, Alex and Kirillov, Alexander},
  journal=NIPS,
  volume={34},
  pages={17864--17875},
  year={2021}
}

@inproceedings{clark2023,
  title={{Where We Are and What We're Looking At: Query Based Worldwide Image Geo-Localization Using Hierarchies and Scenes}},
  author={Clark, Brandon and Kerrigan, Alec and Kulkarni, Parth Parag and Cepeda, Vicente Vivanco and Shah, Mubarak},
  booktitle=CVPR,
  pages={23182-23190},
  year={2023}
}

@inproceedings{pramanick2022,
  title={{Where in the World is this Image? Transformer-based Geo-localization in the Wild}},
  author={Pramanick, Shraman and Nowara, Ewa M and Gleason, Joshua and Castillo, Carlos D and Chellappa, Rama},
  booktitle=ECCV,
  pages={196-215},
  year={2022}
}

@inproceedings{muller2018,
  title={{Geolocation Estimation of Photos using a Hierarchical Model and Scene Classification}},
  author={Müller-Budack, Eric and Pustu-Iren, Kader and Ewerth, Ralph},
  booktitle=ECCV,
  pages={563-579},
  year={2018}
}

@inproceedings{seo2018,
  title={{CPlaNet: Enhancing Image Geolocalization by Combinatorial Partitioning of Maps}},
  author={Seo, Paul Hongsuck and Weyand, Tobias and Sim, Jack and Han, Bohyung},
  booktitle=ECCV,
  pages={536-551},
  year={2018}
}

@inproceedings{weyand2016,
  title={{PlaNet - Photo Geolocation with Convolutional Neural Networks}},
  author={Weyand, Tobias and Kostrikov, Ilya and Philbin, James},
  booktitle=ECCV,
  pages={37-55},
  year={2016}
}

@inproceedings{vo2017,
  title={{Revisiting IM2GPS in the Deep Learning Era}},
  author={Vo, Nam and Jacobs, Nathan and Hays, James},
  booktitle=ICCV,
  pages={2621-2630},
  year={2017}
}

@inproceedings{zhu2022,
  title={{TransGeo: Transformer Is All You Need for Cross-view Image Geo-localization}},
  author={Zhu, Sijie and Shah, Mubarak and Chen, Chen},
  booktitle=CVPR,
  pages={1162-1171},
  year={2022}
}

@article{lin2022,
  title={{Joint Representation Learning and Keypoint Detection for Cross-View Geo-Localization}},
  author={Lin, Jinliang and Zheng, Zhedong and Zhong, Zhun and Luo, Zhiming and Li, Shaozi and Yang, Yi and Sebe, Nicu},
  journal={IEEE Transactions on Image Processing},
  volume={31},
  pages={3780-3792},
  year={2022},
  publisher={IEEE}
}

@inproceedings{zhang2023,
  title={{Cross-View Geo-Localization via Learning Disentangled Geometric Layout Correspondence}},
  author={Zhang, Xiaohan and Li, Xingyu and Sultani, Waqas and Zhou, Yi and Wshah, Safwan},
  booktitle={Proceedings of the AAAI Conference on Artificial Intelligence},
  volume={37},
  number={3},
  pages={3480-3488},
  year={2023}
}

@inproceedings{theiner2022,
  title={{Interpretable Semantic Photo Geolocation}},
  author={Theiner, Jonas and M{\"u}ller-Budack, Eric and Ewerth, Ralph},
  booktitle={Proceedings of the IEEE/CVF Winter Conference on Applications of Computer Vision},
  pages={750-760},
  year={2022}
}

@article{zhang2023text,
  title={{A Survey of Controllable Text Generation Using Transformer-based Pre-trained Language Models}},
  author={Zhang, Hanqing and Song, Haolin and Li, Shaoyu and Zhou, Ming and Song, Dawei},
  journal={ACM Computing Surveys},
  volume={56},
  number={3},
  pages={1-37},
  year={2023},
  publisher={ACM New York, NY}
}

@inproceedings{shao2023,
  title={{Prompting Large Language Models with Answer Heuristics for Knowledge-based Visual Question Answering}},
  author={Shao, Zhenwei and Yu, Zhou and Wang, Meng and Yu, Jun},
  booktitle=CVPR,
  pages={14974-14983},
  year={2023}
}

@article{wei2022,
  title={{Chain-of-Thought Prompting Elicits Reasoning in Large Language Models}},
  author={Wei, Jason and Wang, Xuezhi and Schuurmans, Dale and Bosma, Maarten and Xia, Fei and Chi, Ed and Le, Quoc V and Zhou, Denny and others},
  journal=NIPS,
  volume={35},
  pages={24824-24837},
  year={2022}
}

@article{yao2023tree,
  title={{Tree of Thoughts: Deliberate Problem Solving with Large Language Models}},
  author={Yao, Shunyu and Yu, Dian and Zhao, Jeffrey and Shafran, Izhak and Griffiths, Tom and Cao, Yuan and Narasimhan, Karthik},
  journal=NIPS,
  volume={36},
  year={2024}
}

@inproceedings{dai2023can,
  title={{Why Can GPT Learn In-Context? Language Models Implicitly Perform Gradient Descent as Meta-Optimizers}},
  author={Dai, Damai and Sun, Yutao and Dong, Li and Hao, Yaru and Ma, Shuming and Sui, Zhifang and Wei, Furu},
  booktitle={Findings of the Association for Computational Linguistics: ACL 2023},
  pages={4005-4019},
  year={2023}
}

@article{liu2023llava,
  title={{Visual Instruction Tuning}},
  author={Liu, Haotian and Li, Chunyuan and Wu, Qingyang and Lee, Yong Jae},
  journal=NIPS,
  volume={36},
  year={2024}
}

@article{bai2023qwen,
  title={{Qwen-VL: A Versatile Vision-Language Model for Understanding, Localization, Text Reading, and Beyond}},
  author={Bai, Jinze and Bai, Shuai and Yang, Shusheng and Wang, Shijie and Tan, Sinan and Wang, Peng and Lin, Junyang and Zhou, Chang and Zhou, Jingren},
  journal={arXiv preprint arXiv:2308.12966},
  year={2023}
}

@inproceedings{kenton2019,
  title={{BERT: Pre-training of Deep Bidirectional Transformers for Language Understanding}}, 
  author={Kenton, Jacob Devlin Ming-Wei Chang and Toutanova, Lee Kristina},
  booktitle={Proceedings of the 2019 Conference of the North American Chapter of the Association for Computational Linguistics: Human Language Technologies, Volume 1 (Long and Short Papers)},
  pages={4171–4186},
  year={2019}
}

@inproceedings{reimers2019,
  title={{Sentence-BERT: Sentence Embeddings using Siamese BERT-Networks}},
  author={Reimers, Nils and Gurevych, Iryna},
  booktitle={Proceedings of the 2019 Conference on Empirical Methods in Natural Language Processing and the 9th International Joint Conference on Natural Language Processing},
  pages={3980-3990},
  year={2019}
}

@inproceedings{dosovitskiy2020image,
  title={{An Image is Worth 16x16 Words: Transformers for Image Recognition at Scale}},
  author={Dosovitskiy, Alexey and Beyer, Lucas and Kolesnikov, Alexander and Weissenborn, Dirk and Zhai, Xiaohua and Unterthiner, Thomas and Dehghani, Mostafa and Minderer, Matthias and Heigold, Georg and Gelly, Sylvain and others},
  booktitle=ICLR,
  year={2021}
}

@article{haas2023learning,
  title={{Learning Generalized Zero-Shot Learners for Open-Domain Image Geolocalization}},
  author={Haas, Lukas and Alberti, Silas and Skreta, Michal},
  journal={arXiv preprint arXiv:2302.00275},
  year={2023}
}

@article{gpt4v,
  title={{GPT-4 Technical Report}},
  author={Achiam, Josh and Adler, Steven and Agarwal, Sandhini and Ahmad, Lama and Akkaya, Ilge and Aleman, Florencia Leoni and Almeida, Diogo and Altenschmidt, Janko and Altman, Sam and Anadkat, Shyamal and others},
  journal={arXiv preprint arXiv:2303.08774},
  year={2023}
}

@inproceedings{hu2021lora,
  title={{LoRA: Low-Rank Adaptation of Large Language Models}},
  author={Hu, Edward J and Shen, Yelong and Wallis, Phillip and Allen-Zhu, Zeyuan and Li, Yuanzhi and Wang, Shean and Wang, Lu and Chen, Weizhu},
  booktitle=ICLR,
  year={2022}
}

@inproceedings{rao2023retrieval,
  title={{Retrieval-based Knowledge Augmented Vision Language Pre-training}},
  author={Rao, Jiahua and Shan, Zifei and Liu, Longpo and Zhou, Yao and Yang, Yuedong},
  booktitle={Proceedings of the 31st ACM International Conference on Multimedia},
  pages={5399-5409},
  year={2023}
}

@article{li2021caption,
  title={{Caption Generation From Road Images for Traffic Scene Modeling}},
  author={Li, Yaochen and Wu, Chuan and Li, Ling and Liu, Yuehu and Zhu, Jihua},
  journal={IEEE Transactions on Intelligent Transportation Systems},
  volume={23},
  number={7},
  pages={7805-7816},
  year={2021},
  publisher={IEEE}
}

@article{ying2024mmt,
  title={{MMT-Bench: A Comprehensive Multimodal Benchmark for Evaluating Large Vision-Language Models Towards Multitask AGI}},
  author={Ying, Kaining and Meng, Fanqing and Wang, Jin and Li, Zhiqian and Lin, Han and Yang, Yue and Zhang, Hao and Zhang, Wenbo and Lin, Yuqi and Liu, Shuo and others},
  journal={arXiv preprint arXiv:2404.16006},
  year={2024}
}

@article{xu2023lvlm,
  title={{LVLM-eHub: A Comprehensive Evaluation Benchmark for Large Vision-Language Models}},
  author={Xu, Peng and Shao, Wenqi and Zhang, Kaipeng and Gao, Peng and Liu, Shuo and Lei, Meng and Meng, Fanqing and Huang, Siyuan and Qiao, Yu and Luo, Ping},
  journal={arXiv preprint arXiv:2306.09265},
  year={2023}
}

@inproceedings{hays2008im2gps,
  title={{IM2GPS: estimating geographic information from a single image}},
  author={Hays, James and Efros, Alexei A},
  booktitle=CVPR,
  pages={1--8},
  year={2008},
  organization={IEEE}
}

@inproceedings{luo2022g3,
  title={G3: Geolocation via Guidebook Grounding},
  author={Luo, Grace and Biamby, Giscard and Darrell, Trevor and Fried, Daniel and Rohrbach, Anna},
  booktitle={Findings of the Association for Computational Linguistics: EMNLP 2022},
  pages={5841--5853},
  year={2022}
}
\bibliographystyle{icml2024}

\newpage
\appendix
\onecolumn
\section{Implementation Details}

Table~\ref{tab:settings_details} and Table~\ref{tab:training_details} present the hyper-parameter settings and training details for the models. We conducted training and testing on Nvidia A800 (80G), with CUDA 12.1, PyTorch 2.0.0, and Transformers 4.33.0. 

\begin{table}[h]
\footnotesize
\caption{The hyper-parameter settings of the proposed \emph{GeoReasoner}.} 
\label{tab:settings_details} 
\begin{center} 
\begin{tabular}{cc} 
\toprule 
Hyper Params & Value \\
\midrule 
Learning Rate & 1e-5 \\
Total Batch Size & 64 \\
Weight Decay & 0.1 \\
Warmup Ratio & 0.01 \\
Optimizer & AdamW \\
Adam Beta1 & 0.9 \\
Adam Beta2 & 0.95 \\
LR Scheduler & cosine \\
Model Max Length & 2048 \\
\bottomrule 
\end{tabular} 
\end{center} 
\end{table}

\vspace{-5mm}

\begin{table}[h]
\footnotesize
\caption{The training details of the proposed \emph{GeoReasoner}.} 
\label{tab:training_details} 
\begin{center} 
\begin{tabular}{lccccc} 
\toprule 
& \multirow{2}*{Training Speed} & \multirow{2}*{Inference Latency} & \multicolumn{2}{c}{Num of Params} & \multirow{2}*{Flops} \\
&  &  & Base Model & LoRA &  \\
\midrule 
LoRA1 (reason) & 0.41 sample/s & 1.560s & 9.6B & 112.19M & 71.9B  \\
LoRA2 (location) & 0.63 sample/s & 0.894s & 9.6B & 112.19M & 71.9B \\

\bottomrule 
\end{tabular} 
\end{center} 
\end{table}

\section{Additional Qualitative Results}
\begin{figure}[h]
  \begin{center}
  \centerline{\includegraphics[width=0.99\columnwidth]{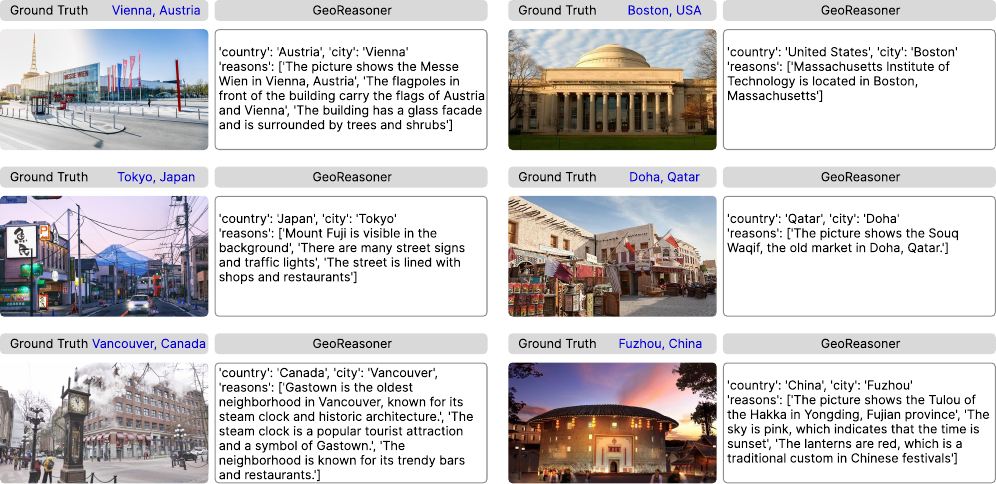}}
  \vspace{-3mm}
  \caption{Additional qualitative results from the proposed \emph{GeoReasoner}.}
  \vspace{-6mm}
  \label{fig:cases}
  \end{center}
\end{figure}

Additionally, we present the results of the \emph{GeoReasoner} on alternative street-view images, depicted in Figure~\ref{fig:cases}. 
Each street view image is annotated with the ground truth geographic location, along with the inference results from \emph{GeoReasoner}. 
It can provide geographical predictions accompanied by reasonable explanations.


\end{document}